\documentclass[runningheads,a4paper]{llncs}
\usepackage{graphicx}
\usepackage{floatrow}
\usepackage[english]{babel}
\usepackage{subfig}
\usepackage{amssymb}
\usepackage{amsmath}
\usepackage[T1]{fontenc}
\usepackage[utf8]{inputenc}
\usepackage{babel}
\usepackage[font=small,labelfont=bf]{caption}
\usepackage[ruled,vlined,linesnumbered,noresetcount]{algorithm2e}
\usepackage{authblk}
\usepackage[T1]{fontenc}
\usepackage[utf8]{inputenc}
\let\origtau\tau 
\renewcommand{\tau}{\scalebox{1.44}{$\origtau$}}
\newcommand{\squeezeup}{\vspace{-2.5mm}}
\usepackage{url}
\newcommand{\keywords}[1]{\par\addvspace\baselineskip
\noindent\keywordname\enspace\ignorespaces#1}

\begin{document}
\mainmatter  
\title{A Reinforcement Learning Approach for Sequential Spatial Transformer Networks}
\titlerunning{Sequential Spatial Transformer Networks}
\author{Fatemeh Azimi\textsuperscript{1,2} \and Federico Raue\textsuperscript{1} \and J\"orn Hees\textsuperscript{1} \and Andreas Dengel\textsuperscript{1,2}}
\authorrunning{Azimi \and Raue \and Hees \and Dengel}
\institute{\textsuperscript{1} TU Kaiserslautern, Germany\\
\textsuperscript{2}Smart Data and Knowledge Services, German Research Center for Artificial Intelligence (DFKI), Germany\\
\texttt{\{fatemeh.azimi, federico.raue, joern.hees, andreas.dengel\}@dfki.de}\
}
\toctitle{?}
\tocauthor{?}
\maketitle
\begin{abstract}
Spatial Transformer Networks (STN) can generate geometric transformations which modify input images to improve classifier's performance.
In this work, we combine the idea of STN with Reinforcement Learning (RL).
To this end, we break the affine transformation down into a sequence of simple and discrete transformations.
We formulate the task as a Markovian Decision Process (MDP) and use RL to solve this sequential decision making problem.
STN architectures learn the transformation parameters by minimizing the classification error and backpropagating the gradients through a sub-differentiable sampling module.
In our method, we are not bound to differentiability of the sampling modules.
Moreover, we have freedom in designing the objective rather than only minimizing the error;
e.g., we can directly set the target as maximizing the accuracy.
We design multiple experiments to verify the effectiveness of our method using cluttered MNIST and Fashion-MNIST datasets and show that our method outperforms STN with proper definition of MDP components.
\end{abstract}
\section{Introduction}
Invariance against different transformations is crucial in many tasks such as image classification and object detection.
Previous works have addressed this challenge, from early work on feature descriptors \cite{lowe2004distinctive} to modeling geometric transformations \cite{dai2017deformable}.
It is also very beneficial if the network can detect the important content in the image and distinguish it from the rest \cite{redmon2016you}.
To this end, there are different approaches such as searching through region proposals in object detection \cite{girshick2015fast,ren2015faster}, and using various attention 
mechanisms for both classification and detection tasks \cite{mnih2014recurrent,ba2014multiple,sermanet2014attention,bueno2017hierarchical}.

With recent advance in deep learning, there has been a breakthrough in various areas of Computer Vision mainly caused by the advances in Convolution Networks \cite{lecun1998gradient,krizhevsky2012imagenet}.
Introducing deeper and more complex classification network architectures \cite{he2016deep,szegedy2015going,huang2017densely} has led to achieving high accuracy in challenging datasets such as ImageNet \cite{deng2009imagenet}. 
However, another approach for improving the performance is to simplify the classification by transforming the input image \cite{jaderberg2015spatial}.
Hence, an important question to ask is \textit{"what are the suitable transformations?"}.

In \cite{jaderberg2015spatial}, the authors introduced the STN method for improving the classification accuracy.
In STN, a network is trained to generate parameters of an affine transformation which is applied to the input image.
They showed that this modification simplified the task and improved the performance.
In their work, affine parameters were searched locally by differentiating the classification loss and backpropagating the gradients through a sub-differentiable sampling module.

Similar to STN, we address improving the classifier accuracy by applying an affine transformation to the input.  
Different from their approach, we model the task as a Markovian Decision Process.
We break the affine transformation to a sequence of discrete and simple transformations and use RL to search for a combination of transformations which minimizes the classification error.
This way, the task is simplified to a search problem in discrete search space.
Using RL, we are not dependent on differentiability of different sampling modules and not limited to minimizing the classification loss as the optimization objective.

Since the breakthrough in RL \cite{mnih2013playing}, many works have successfully utilized it for solving different vision problems \cite{zoph2016neural,baker2016designing,liang2017deep,park2018distort}.
Combining RL methodology with deep learning as well as significant improvement in RL algorithms \cite{schulman2017proximal,wu2017scalable} has made it a powerful search method for different applications \cite{zoph2016neural,cubuk2018autoaugment}.
Moreover, RL can serve as a learning method which is not dependent on differentiability of the utilized modules \cite{mnih2014recurrent}.
%
For example, \cite{yu2018crafting} adapted an RL solution for Image Restoration (IR), in which the goal is maximizing the Peak Signal to Noise Ratio (PSNR).
For this task, they provide a set of IR tools and use RL to search for the optimal combination of applying these tools, aiming to maximize PSNR.
In another application, Bahdanau at al. adapted RL for language sequence prediction \cite{bahdanau2016actor}.
In RL framework, one can design a reward for different objectives;
they used this characteristic to directly search for a sequence which maximizes the test time metrics such as BLEU score. 

To sum up, we formulate the transformation task as a sequential decision-making problem,
in which instead of finding a one-step transformation, the model searches for a combination of discrete transformations to improve the performance.
We use RL for solving the search problem and apply both Policy Gradient and Actor-Critic algorithms \cite{sutton2000policy,sutton1984temporal}.
We experiment with different reward designs including maximizing classification accuracy and minimizing the classification loss.
In the following, we provide related work and the required background for our approach. 
Afterwards, we explain our method followed by experiments and an ablation study.
%
%
%
\section{Related Work}
\squeezeup
Our work is mainly related to STN model and the RL algorithms that we utilize for solving the sequential transformation task.
In this part, we focus on explaining the main ideas of STN approach as well as the required background about RL algorithms.
%
\squeezeup
\subsection{Spatial Transformer Network (STN)}
\begin{figure}
    \centering
    \includegraphics[width=10cm]{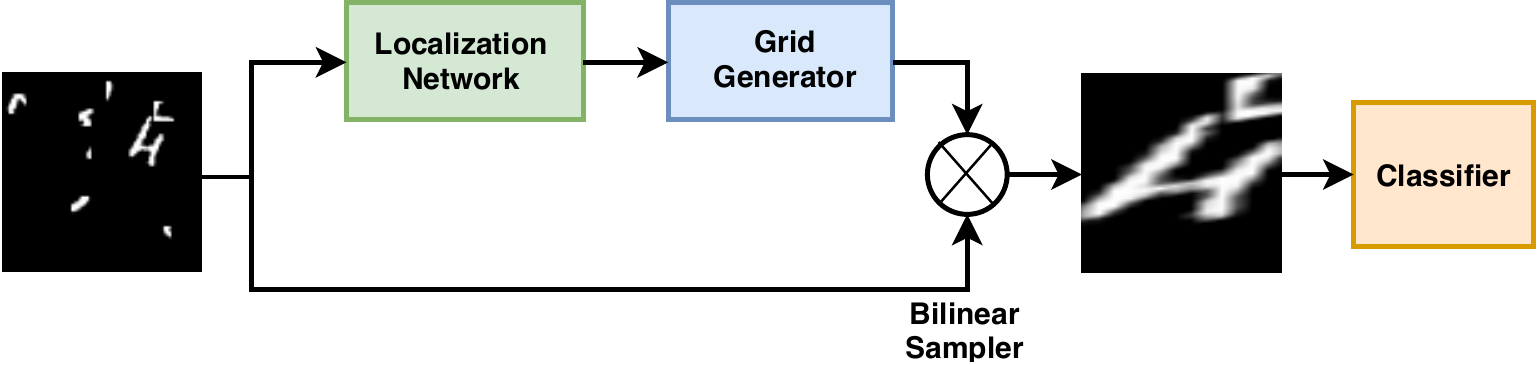}
    \caption{An overview of the components in STN architecture. The localization network generates the parameters of an affine transformation. The grid generator together with the sampler module generates the transformed image.}
    \label{fig:stn}
\end{figure}

In STN architecture \cite{jaderberg2015spatial}, the model learns a geometric transformation and modifies the input image to minimize the classification error.
Although it is possible to use different transformations, here we focus only on affine transformation.
%
The main components of an STN are the \textit{localization network}, the \textit{grid generator} and the \textit{sampler module}.
Figure \ref{fig:stn} shows an overall view of STN architecture.
The \textit{localization network} takes the input image and generates the affine transformation parameters.
The \textit{grid generator} computes the location of each output pixel in the input image.
To warp the input image based on the estimated transformation, each pixel in the output should be computed using a sampling kernel applied to the input image.
The \textit{sampler} uses the grid generator output and the bilinear sampling kernel to generate output pixels from the input image:
\begin{equation}
    V_{ij} = \sum_{n}^{H}\sum_{m}^{W}U_{nm}max(0,1-|x_{i} - m|)max(0,1-|y_{j} - n|)\, \forall i,j\in [1\,...\,H],[1\,...\,W],
    \label{eq:kernel}
\end{equation}
where \emph{H} and \emph{W} are the height and width of the image respectively, and \emph{V} and \emph{U} are the corresponding pixel values in the output and input image.
The coordinate $(x_{i},y_{i})$ is the location in the input where the sampling kernel is applied. 
The sampling module is differentiable within the local neighborhood, as can be seen in Equation \ref{eq:bi-grad}. 
\begin{equation}
    \frac{\delta V_{ij}}{\delta x_{i}} = U_{nm}max(0,1-|y_{j}-n|)\begin{cases}
    0 & \text{if } |m-x_{i}|\geq{1}\\
    1 & \text{if } m\geq{x_{i}}\\
    -1 & \text{if } m\leq{x_{i}}
    \end{cases}
    \label{eq:bi-grad}
\end{equation}
and similarly for $\frac{\delta V_{ij}}{\delta y_{j}}$.
Therefore, the parameters of the localization network are gradually updated using backpropagation through the classification loss within a local window.
For more information, please refer to the original paper \cite{jaderberg2015spatial}.
\subsection{Reinforcement Learning (RL)}
\label{seq:RL}
The main components in an RL framework are the State Space (\textbf{S}), the Action Space (\textbf{A}), and the Reward Signal (\textbf{R}) \cite{sutton2018reinforcement}.
Additionally, an \textit{episode} refers to a sequence of state-action transitions from the initial state until the final state.
An important consideration in defining \textbf{S} is the Markovian assumption; 
it implies that selecting an action only requires information from the current state.
Having this framework, a network (or an agent) is trained to learn picking the right action (or a policy) at every state in the episode.
In training an RL agent, the objective is maximizing the expected total reward at the end of each episode:
\begin{equation}
    J(\theta) = E_{\origtau\sim \pi_{\theta}}[r(\origtau)],
    \label{eq:objective}
\end{equation}
where $r=\sum_{t=0}^{t=T} R_{t}$, \textit{T} is the episode length, and $\origtau$ is an episode sampled from policy $\pi_{\theta}$.

RL provides two main training algorithms: Policy Gradient (PG) and Q-learning.
Additionally, there are Actor-Critic (AC) algorithms, which combine PG and Q-learning to merge the advantages of both algorithms.
In PG algorithms, the policy is often approximated using a neural network which is trained by maximizing the objective in Equation \ref{eq:objective}, using gradient ascent.
Using backpropagation for maximizing this objective leads to the update rule below, with $\alpha$ as the learning rate.
\begin{equation}
    \theta_{new} = \theta_{old} + \alpha \nabla_{\theta}J(\theta)
\label{eq:pg1_1}
\end{equation}
\begin{equation}
    \nabla_{\theta}J(\theta) = E_{\origtau\sim \pi_{\theta}}[\nabla_{\theta}(\log{\pi_{\theta}(\origtau)})r(\origtau)]
    \label{eq:pg1_2}
\end{equation}
In Q-learning the optimal policy is found by estimating the Q function.
The Q function approximates the expected total reward from the current state for each possible action: 
\begin{equation}
    Q^{\pi}(s_{t},a_{t}) = E_{\pi_{\theta}}\left[\sum_{t'=t}^{T}r(s_{t'},a_{t'}|s_{t},a_{t})\right]
    \label{eq:Q}
\end{equation}
Therefore, one can find the optimal policy by always selecting the action leading to higher Q value.

PG methods are simple and effective, but they suffer from high variance.
In AC, this issue is addressed by subtracting a baseline from $r(\origtau)$ in Equation \ref{eq:pg1_2}, and shifting it around the zero mean.
More precisely, the baseline in AC is the expected value of the Q function as
\begin{equation}
    V^{\pi}(s_{t}) = E_{a_{t} \sim \pi_{\theta}}\left[Q^{\pi}(s_{t},a_{t})\right] 
    \label{eq:V}
\end{equation}
This value is also approximated using another neural network.
The value network is supposed to estimate the total reward from time-step \emph{t} onward.
Therefore, the sum of rewards from \emph{t} to the end of the episode ($\sum_{t=t'}^{T}r(t)$) can serve as the ground-truth for training the value network using a proper loss function.
More details can be found in \cite{sutton2018reinforcement}.
In this paper, we experiment with both PG and AC algorithms.
\section{Sequential Spatial Transformer Network (SSTN)}
Our goal is to learn a sequence of image transformations $T = T_{n} \cdot T_{n-1} \cdot ... \cdot T_{0}$, which is applied to the input image and helps the classifier achieve a better performance.
There are different image adjustments including geometric transformations and filtering methods.
In this paper, we only consider affine transformations.

\begin{figure}
    \centering
    \includegraphics[width=\textwidth]{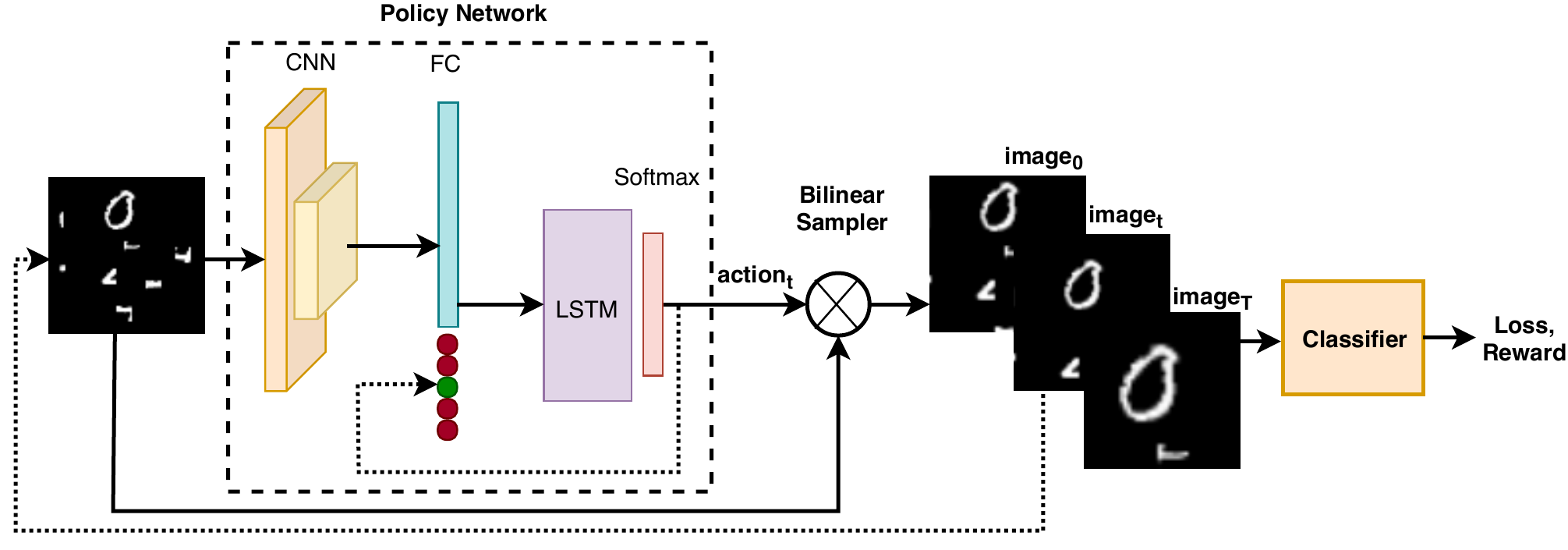}
    \caption{SSTN architecture for finding the sequential affine transformation $T = T_{n} \cdot T_{n-1} \cdot ... \cdot T_{0}$.
    We compare the performance of different policy architectures with and without using LSTM.
    The last layer of the policy network is a softmax which outputs the probability distribution of the actions.
    When using LSTM, the one-hot encoded action from the previous time-step is merged into the feature map of $image_{t}$.
    At each time-step an action corresponding to transformation $T_{i}$ is sampled from the policy and applied to the image. 
    }
    \label{fig:SSTN}
\end{figure}

In this paper, we decompose the affine transformation into a sequence of specific and discrete transformations instead of applying it in one step as in \cite{jaderberg2015spatial}.
We formulate the problem of finding the affine parameters as an MDP and aim to learn picking the right transformation at every time-step of the sequence. 
Figure \ref{fig:SSTN} shows our proposed architecture.
\squeezeup
\squeezeup
\subsection{MDP Framework for SSTN}
\label{sec:method}
As mentioned in Section \ref{seq:RL}, the main parts in an RL framework are \textbf{S}, \textbf{A}, and \textbf{R}.
In this section, we elaborate on these elements in formulating our task.
\squeezeup
\subsubsection{State Space:}
We consider two state space definitions and experiment with both of them.
First, we define a state as the transformed image at step \emph{t} ($image_{t}$ in Figure \ref{fig:SSTN}).
Second, we define the state as a combination of the current transformed image ($image_{t}$) and the previous action ($a_{t-1}$).
We merge the one-hot encoded action from the last time-step into the state as: $s_{t}=(image_{t}, action_{t-1})$.
To keep track of the order of sampled actions in different time-steps, the model utilizes an LSTM module \cite{hochreiter1997long}, which is a recurrent neural network with gate functions to avoid vanishing gradient problem.
This formulation is closer to the Markovian assumption as the information from the past actions helps the network to learn the proper order of applying the transformations.
\squeezeup
\subsubsection{Action Space:}
Every action $action_{t}$ is a specific transformation sampled from the policy which is applied at time-step \emph{t} and slightly transforms the image and the state.
In order to construct an affine transformation, we define the action space as $A=\{Translation, Rotation, Scale, Identity\}$.
The episode length is fixed to \textit{T} for all images, and having the \textit{Identity} transformation allows for stopping the process for individual images before reaching \textit{T}.
Having a fixed episode length allows us to train our model in mini-batches.
\squeezeup
\subsubsection{Reward:}
The agent learns the task while maximizing the reward;
therefore, the reward definition has to enfold the objective of the task.
An intuitive reward definition would be based on classification accuracy, since the goal is achieving higher accuracy.
Accordingly, we give a discrete reward of \textbf{+1} when a label prediction changes from false to correct as the result of applying an action, and \textbf{-1} for the opposite case;
other cases get \textbf{0} reward:
\begin{equation}
    r_{1} = \begin{cases}
    \boldsymbol{1} & \text{if } (pred_{t-1} \ne label \land pred_{t}=label)\\
    \boldsymbol{-1} & \text{if } (pred_{t-1}=label \land pred_{t} \ne label)\\
    \boldsymbol{0} & \text{otherwise} \\
    \end{cases}
    \label{eq:acc_rew}
\end{equation}
Here $pred_{t-1}$ and $pred_{t}$ are predicted labels before and after applying the action at time-step \emph{t} $a_{t}$.

Moreover, we can address maximizing the accuracy by minimizing the classification loss similar to \cite{jaderberg2015spatial}.
This way the reward design is simply the negated loss:
\begin{equation}
    r_{2} = -loss
\end{equation}
In this case, the reward is always negative as loss is a positive value;
therefore, the maximum expected reward would be zero.
It means that that the model tries to learn a policy which pushes the classification loss toward zero.

Additionally, we can consider the reward as the loss difference between consecutive time-steps:
\begin{equation}
    r_{3} = loss_{t-1} - loss_{t},
\label{eq:relative_rew}
\end{equation}
where \emph{t} is the time-step.
With this reward definition, the model tries to maximize the difference in loss values between every two following steps, toward a positive reward.
In other words, the model tries to pick an action which results in a smaller loss value compared to the previous time-step.
\subsection{Training}
Having \textbf{S}, \textbf{A}, and \textbf{R} defined, we use the algorithms introduced in Section \ref{seq:RL} to learn combining a set of discrete transformations for improving the classifier performance.
First, we use PG algorithm mainly due to its effectiveness and simplicity.
Then we extend our implementation to AC algorithm. 
More details about AC training algorithm for one epoch is presented in Algorithm \ref{al1}. 
PG algorithm is similar, but it does not include the critic network and baseline reduction.\\

\begin{algorithm}[H]
\SetAlgoLined
\SetKwInOut{Input}{Input}
\SetKwInOut{Output}{Output}
    \Input{Images and classification labels}
    \Output{Classifier and sequential spatial transformer policy}
    \textit{Initialize classifier $(\psi)$, actor $(\theta)$, and critic $(\phi)$ networks}  \\
 \For{Image = 1 : N}{
    $initialize \ state \ \boldsymbol{s}, \ t=0, \ RewardValueList = [empty]$\\
    \While{t $<$ episode-length}{
    $\pi(a_{t}|s_{t}) = actor(s_{t}$)\\
    $sample \ the \ action: \ a_{t}\sim \pi(a_{t}|s_{t})$\\
    $image_{t+1} = \boldsymbol{apply\_ction}$($image_{t}, a_{t}$)\\
    $\boldsymbol{s_{t+1}} = (image_{t+1}, \boldsymbol{one\_hot}(a_{t}))$\\
    $predictions_{t}, loss_{t} = \boldsymbol{classify}(image_{t}, labels)$\\
    $reward_{t} = \boldsymbol{compute\_reward}$\footnote{Depending on the used reward definition (Equations \ref{eq:acc_rew} to \ref{eq:relative_rew}) }\\
    $value_{t} = critic(s_{t})$ \\
    $append \ (reward_{t},value_{t})\  to \  RewardValueList$\\
    $t \ += \ 1$
    }
  \textbf{Update parameters:}\\
   $\Delta\phi = \frac{d}{d\phi}MSE\_loss(v_{t}, \sum_{t=t'}^{t=T}\gamma^{T-t}r_{t})$\footnote{$\gamma$ is a discount factor, here set to 0.98}\\
   $\Delta\theta = \frac{d}{d\theta }E[log\pi_{\theta}(\sum_{t=t'}^{t=T}\gamma^{T-t}r_{t}-v_{t})]$\\
   $\Delta\psi = \frac{d}{d\psi}CrossEntropy\_loss(predicted\_labels, labels)$\\
   $\boldsymbol{Empty}(RewardValueList)$
 }
 \caption{Actor-Critic Training Algorithm for SSTN}
 \label{al1}
\end{algorithm}
\section{Experiments}
\squeezeup
In this section, we present the experimental setup for testing the performance of our method in improving the classification accuracy by applying a sequence of discrete transformations to the input image.
We proceed with a discussion on results and an ablation study on the impact of reward design and episode length.
\squeezeup
\squeezeup
\squeezeup
\subsubsection{Dataset:}
We evaluate our method using cluttered MNIST \cite{mnih2014recurrent} and cluttered Fashion-MNIST datasets. 
cluttered MNIST has been used by several works to demonstrate visual attention~\cite{gregor2015draw,mnih2014recurrent}. 
We followed the same procedure for generating cluttered Fashion-MNIST which includes ten clothing categories.
We generate $80\times80$ grayscale images using the publicly available code\footnote{https://github.com/deepmind/mnist-cluttered}.
The generated images are covered by clutter and the main content is located at a random location within the image boundaries.
Both datasets include 500K training and 100K test images.
\squeezeup
\subsection{Network Architecture and Experiment Description}
\squeezeup
Our action space includes 10 transformations, including $\pm4\ pixels$ translation in \emph{x} and \emph{y} direction,
scaling of 0.8 in \emph{x}, \emph{y}, and \emph{xy} direction (since transformation is applied in backward mapping manner, $scale<1$ has zoom-in effect), rotation of $\pm10$ degrees and \emph{Identity} transformation. 
In Section \ref{seq:ablation}, we show a comparison of the classification accuracies using different reward definitions.
Here we present results using reward definition in Equation \ref{eq:relative_rew}.

We evaluate our model in the following settings.
First, we take a 2-layer fully connected network as the classifier (referred as MLP in the following).
The reason for choosing this simple classifier is to examine the improvement in classification accuracy based on only image transformations.
Keeping the classifier architecture unchanged, we experiment with different policy architectures including LeNet and LeNet combined with LSTM.
We also experiment with both PG and AC training algorithms.

For comparison, we implement our own version of the STN model and train it on our dataset (the size of cluttered MNIST images is $60\times60$ in STN paper).
For the localization network in Figure \ref{fig:stn}, we use the same LeNet architecture as in policy network.
In STN paper \cite{jaderberg2015spatial}, they used SGD as optimizer; however, we reached better results using Adam optimizer and we report the best observed performance.
\begin{center}
\begin{tabular}{l*{6}{c}r}
\textbf{Method} & \textbf{CMNIST(\%)} &\textbf{CFMNIST(\%)}\\ \hline 
     MLP classifier &  54.01 &30.74\\ 
     MLP classifier with STN  & 94.49& 62.54 \\ \hline
     LeNet policy using PG algorithm & 91.04 &58.70 \\ 
     LeNet+LSTM policy using PG algorithm & 95.88&70.27\\ 
     LeNet+LSTM policy using AC algorithm  & \textbf{96.83}& \textbf{71.61}\\ 
\end{tabular}
\captionof{table}{Experiments with MLP classifier and STN as the baseline, followed by the SSTN approach results.
  For all the experiments the classifier architecture is the same.
  The episode length for SSTN is set to 40.
   Our experiments cover PG algorithm with different architectures as well as applying AC algorithm on the best architecture.
   CMNIST and CFMNIST columns are classification accuracy for cluttered MNIST and Fashion-MNIST datasets, respectively.
   We observe that with proper definition of the state space, our approach outperforms the STN method.}
   \label{tab:res1}
\end{center}
The utilized LeNet architecture consists of two convolution layers with 32 and 64 kernels, followed by max pooling and two fully-connected layers with ReLU non-linearity.
In policy network, the last layer is a softmax which generates the action probabilities.
The actions are sampled from a \textit{Categorical} distribution fitted to the softmax output.
For all experiments we use Adam optimizer with learning rate of $1e-4$ and the episode length is 40.
In the next experiment setup, we change the classifier to LeNet and repeat similar experiments.
\begin{center}
  \begin{tabular}{ l*{6}{c}r}
     \textbf{Method} & \textbf{CMNIST(\%)}&\textbf{CFMNIST(\%)}\\ \hline 
     LeNet classifier  & 95.94 &72.40\\ 
     LeNet classifier with STN & 97.72&77.38 \\ \hline
     LeNet policy using PG algorithm & 95.82 & 74.59\\
     LeNet+LSTM policy using PG algorithm & 98.23 & 83.16\\
     LeNet+LSTM policy using AC algorithm & \textbf{98.29}&\textbf{83.27}\\ 
  \end{tabular}
  \captionof{table}{Experiment results when using LeNet for the classifier network and the episode length of 40.
  CMNIST and CFMNIST columns are the accuracy results for cluttered MNIST and Fashion-MNIST datasets, respectively.}
   \label{tab:rescnn}
\end{center}
Tables \ref{tab:res1} and \ref{tab:rescnn} show the results of our experiments using MLP and LeNet classifiers.
We note that the impact of both policy network in SSTN and localization network in STN is only to transform the image before feeding it to the classifier and not to increase the power of the classifier.

For the policy network, first we use a LeNet architecture and then combine it with LSTM.
We aim to investigate if considering the state as the current single image satisfies the Markovian assumption.
Based on the experiments with LSTM module, we observe that this is an essential element and the single image does not include all the required information.
The reason is that the RL agent is supposed to learn the sequence of actions constructing the optimal affine transformation;
therefore, it needs the tool for remembering the order of applying actions.
The input to LSTM is the extracted feature map from the current transformed image, concatenated with one-hot encoded previous action.
Finally, we take the best architecture from these experiments and train it with AC algorithm.
In AC algorithm, we use the same network as policy for the critic.
Although it is possible to share weights between the actor and the critic, it is more stable if separate networks are used \cite{bahdanau2016actor}.
As expected, applying AC training algorithm leads to further improvement; 
since it addresses some of the shortcomings in PG as mentioned in Section \ref{seq:RL}.
As results in Table \ref{tab:rescnn} show, the LeNet classifier serves as a strong baseline and achieves high accuracy, especially in cluttered MNIST dataset.
However, we still can get an improvement by modifying the input image before classifying.
\squeezeup
\subsection{Ablation Study}
\label{seq:ablation}
In this section, we present an ablation study on the impact of the reward design and episode length on the performance.
\subsubsection{Reward:}
\begin{figure}
    \centering
    \includegraphics[width=\textwidth]{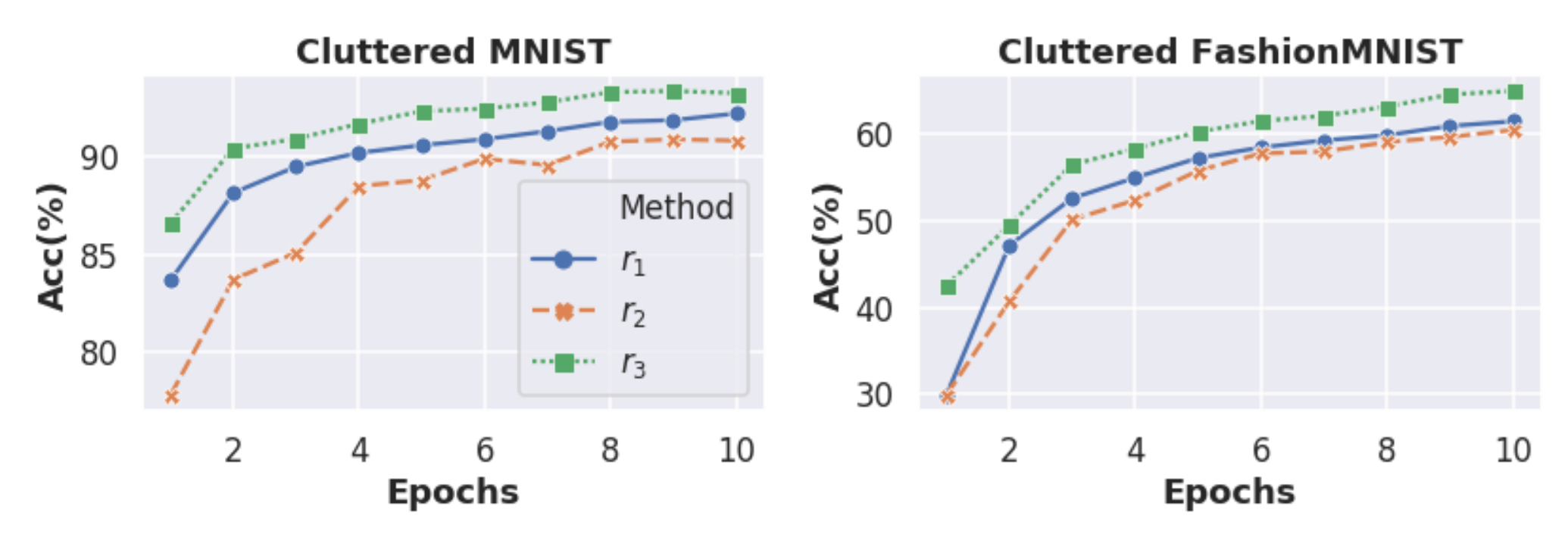}
    \caption{Comparison between three different reward definitions using MLP classifier and LeNet+LSTM policy network using PG algorithm and episode length of 20.} 
    \label{fig:compare_reward} 
\end{figure}

Figure \ref{fig:compare_reward} shows the epoch-accuracy curve for different reward definitions.
Although the performance is close, $r_{3}$ outperforms the others.
We believe this behavior is because $r_{3}$ provides more concise information about the taken action compared to the discrete reward in Equation \ref{eq:acc_rew}.
We observe that the performance using $r_{2}$ is worse than the others.
We argue that in reward $r_{1}$ and $r_{3}$, we consider the change caused by taking the action between every two time-steps;
while, in $r_{2}$ we only consider the loss value at current time-step and not the change.
The results indicate that this formulation incorporates less information compared to the other two.
\subsubsection{Episode Length:}
\begin{figure}
    \centering
    \includegraphics[width=\textwidth]{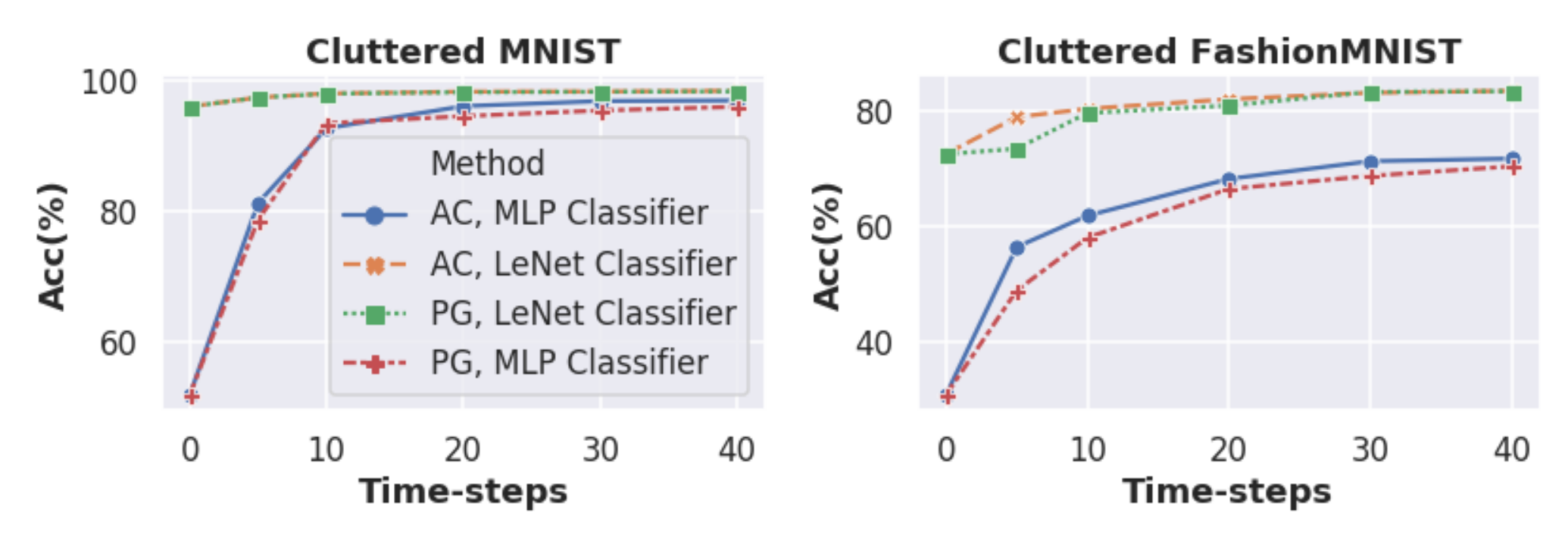}
    \caption{Results for AC and PG algorithms with MLP and LeNet classifiers using different episode lengths.}
    \label{fig:t_compare}
\end{figure}

Another important hyper-parameter is the number of time-steps per episode.
Figure \ref{fig:t_compare} shows the performance of AC and PG algorithms for different time-steps.
As the results illustrate, the accuracy is better for more extended episodes.
However, this can be seen as a trade-off between speed and accuracy.
Another observation is that when using LeNet as classifier, the performance of PG and AC algorithms are very similar.
This indicates that using a stronger classifier decreases the variance in reward signal.
\squeezeup
\section{Conclusion and Future Work}
In this work, we present an extension of the STN model, in which we model the problem as a sequence of discrete transformations.
We formulate finding the affine transformation as a search problem and aim to learn a combination of discrete transformations which improves the classification accuracy.
We use both Policy Gradient and Actor-Critic training algorithms and compare our method with extensive experiments on cluttered MNIST and Fashion-MNIST datasets.
For future work, we would like to extend this work to more complex datasets such as SVHN and PASCAL VOC.
Moreover, we plan to extend our approach to more general transformations beyond geometric alterations, e.g., morphological operations;
this extension can be done by merely extending the action space.
Another exciting direction is adapting this method for other relevant tasks such as detection.
\squeezeup
\squeezeup
\subsubsection{Acknowledgement}
This work was supported by TU Kaiserslautern CS PhD scholarship program, the BMBF project DeFuseNN (Grant 01IW17002), and the NVIDIA AI Lab (NVAIL) program.

\bibliography{ref}
\bibliographystyle{ieeetr}

\end{document}